\documentclass[manuscript,screen,review=false]{acmart}
\AtBeginDocument{%
  \providecommand\BibTeX{{%
    \normalfont B\kern-0.5em{\scshape i\kern-0.25em b}\kern-0.8em\TeX}}}

\copyrightyear{2023}

\acmPrice{15.00}
\acmISBN{978-1-4503-XXXX-X/18/06}

\begin{document}

\title{Robot Patrol: Using Crowdsourcing and Robotic Systems to Provide Indoor Navigation Guidance to The Visually Impaired}


\author{Ike Obi}
\email{obii@purdue.edu}
\affiliation{%
  \institution{Purdue University}
  \streetaddress{401 N Grant Street}
  \city{West Lafayette}
  \state{Indiana}
  \postcode{47907}
  \country{USA}
}

\author{Ruiqi Wang}
\email{wang5357@purdue.edu}
\affiliation{%
  \institution{Purdue University}
  \streetaddress{401 N Grant Street}
  \city{West Lafayette}
  \state{Indiana}
  \postcode{47907}
  \country{USA}
}

\author{Prakash Shukla}
\email{shukla37@purdue.edu}
\affiliation{%
  \institution{Purdue University}
  \streetaddress{401 N Grant Street}
  \city{West Lafayette}
  \state{Indiana}
  \postcode{47907}
  \country{USA}
}

\author{Byung-Cheol Min}
\email{minb@purdue.edu}
\affiliation{%
  \institution{Purdue University}
  \streetaddress{401 N Grant Street}
  \city{West Lafayette}
  \state{Indiana}
  \postcode{47907}
  \country{USA}
}

\renewcommand{\shortauthors}{Obi, et al.}

\begin{abstract}
Indoor navigation is a challenging activity for persons with disabilities, particularly, for those with low vision and visual impairment. Researchers have explored numerous solutions to resolve these challenges; however, several issues remain unsolved, particularly around providing dynamic and contextual information about potential obstacles in indoor environments. In this paper, we developed Robot Patrol, an integrated system that employs a combination of crowdsourcing, computer vision, and robotic frameworks to provide contextual information to the visually impaired to empower them to navigate indoor spaces safely. In particular, the system is designed to provide information to the visually impaired about 1) potential obstacles on the route to their indoor destination, 2) information about indoor events on their route which they may wish to avoid or attend, and 3) any other contextual information that might support them to navigate to their indoor destinations safely and effectively. Findings from the Wizard of Oz experiment of our demo system provide insights into the benefits and limitations of the system. We provide a concise discussion on the implications of our findings.
\end{abstract}




\maketitle

\section{Introduction}


Navigation in an indoor environment requires awareness of continually evolving dynamic and contextual situations like closed elevators, obstacles, cleaning by janitors, crowd levels, noise, wet floors, and other detrimental obstacles. Although sighted users are typically able to navigate these kinds of situations with ease, people with visual impairments struggle a lot in these kinds of environments. Hence, highlighting the need to support them with reliable and near real-time information about obstacles or events on the route to their indoor destination and ultimately to empower them to navigate indoor spaces safely, independently, and effectively. We implicate that combining the strengths of crowdsourcing, computer vision, and robotic systems is a pragmatic approach to generate, validate and circulate contextual information, such as obstacles, public events, in indoor environments. Specifically, the crowd provides the initial contextual information, then robots can patrol the environment periodically to validate and update it via a deep learning-based vision system, providing up-to-date contextual guidance to the visually impaired.

To promote the adoption of the Robot Patrol platform among members of the public, we designed crowdsourcing and gamification approaches. These methods will facilitate the recruitment, engagement, and retention of sighted participants that will provide contextual information about obstacles in the indoor spaces in their proximity. The framework is designed such that users are rewarded with points and badges for every step of their participation in providing information to the systems. They are also to be notified if a visually impaired person uses their information (report) and finds it helpful. Furthermore, the platform also integrated robotic and computer vision frameworks into the system to perform the following tasks: 1) verify the information provided by the participants or members of the public 2) investigate potential obstacles or warnings on the path the blind will pass, and 3) provide feedback of their verification task to the system. The validation and update tasks are very time and labor-consuming. Therefore, using a robot instead of the crowd to do this can 1) reduce human workload and increase interest, etc. 2) provide the ability to continuously update/make the environments more likely to be up-to-date. In all, this approach is designed to provide relevant information to support people with low vision and visual impairment to navigate indoor spaces meaningfully, safely, and effectively. We further conducted a wizard of oz experiment with our system by deploying it in a controlled space to explore its workability, benefits, and limitations. 

Altogether, our exploratory research project contributes insights on the potential implications of employing crowdsourcing, computer vision, and robotic frameworks to provide contextual information and guidance to the visually impaired to empower them to navigate indoor spaces safely. In the next section, we discuss related works and connect our project to previous researches within the field, particularly, in the areas of providing indoor navigation support to the visually impaired.

\section{Related Works}
Various researchers have explored challenges of developing indoor navigation systems for the visually impaired ~\cite{dias2014navpal,kulyukin2006robot,galatas2011eyedog,jeamwatthanachai2019indoor, chen2019crowd,unhelkar2015human,burger2020mobile}. Jeamwatthanachai et. al. ~\cite{jeamwatthanachai2019indoor}, investigated the challenges blind people face while navigating indoors, and surfacing those existing technologies are insufficient. Dias ~\cite{dias2014navpal} explored approaches for empowering blind people to navigate indoors safely and independently. Their project employed robotics, crowdsourcing, path-planning and multimodal interfaces ~\cite{dias2014navpal} only for mapping WiFi locations and landmarks and however, did not extend to providing ‘real-time’ dynamic and contextual information about indoor spaces to the low vision and visually impaired users. Kulyukin ~\cite{kulyukin2006robot,chen2019crowd,unhelkar2015human}, and several other researchers have also demonstrated how robots could be used for wayfinding in the indoor environment for the visually impaired. 

Similarly, prior work by researchers has explored the approach of using crowdsourcing to support the activities of the visually impaired ~\cite{hoonlor2015ucap,xiao2015assistive,rice2013crowdsourcing}. Hoonlor et al. ~\cite{hoonlor2015ucap} designed a mobile app to enable the visually impaired obtain information about consumer products by accessing images in their database crowdsourced by users. Findings from their research revealed that the visually impaired users have positive reviews of the system. Rice et al. ~\cite{rice2013crowdsourcing} identified that the inability to add real-time information to maps affects people with disability and visually impairment. Findings from their research revealed that crowdsourcing framework is a suitable approach for adding information to maps as a way of supporting persons with disability. Xiao et al. ~\cite{xiao2015assistive} also developed an assistive framework for context-aware navigation and as a means of supporting people with visual impairment. Results from their pilot study revealed the suitablity of crowdsourcing framework to providing indoor navigation guidance to the visually impaired. Altogether, the works of these authors highlight that employing crowdsourcing frameworks enables organizations to distribute tasks among a larger network of users and participants than is typically feasible in traditional methods.

\section{Approaches}
The Robot Patrol project has four main sections, including 1) the system (software application), 2) the crowdsourcing framework, 3) the robotic system, 4) the low vision and visually impaired users. The system mediates the connection between the participants, the robot system, and the visually impaired users. The flow map [\ref{fig:image3}] highlights how the information flows from the participants and members of the public to the robot operating system to verify the information provided before being sent back to the system for final delivery to the visually impaired. 

\subsection{System}
\begin{figure}
    \centering
    \includegraphics[width=0.65\linewidth]{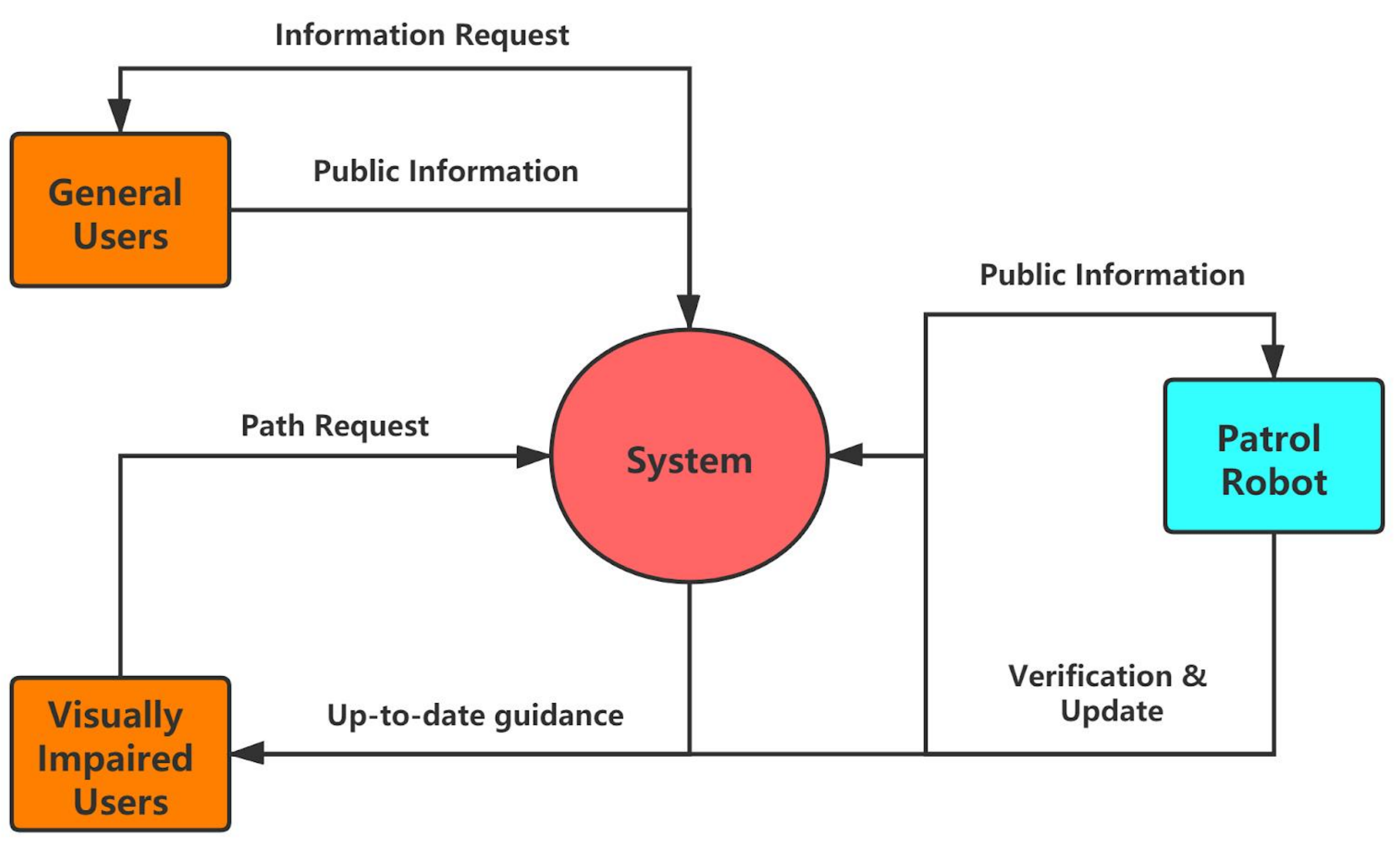}
    \caption{How each component of the platform connects to the central system.} 
    \label{fig:image3}
\end{figure}

\begin{figure}
    \centering
    \includegraphics[width=0.63\linewidth]{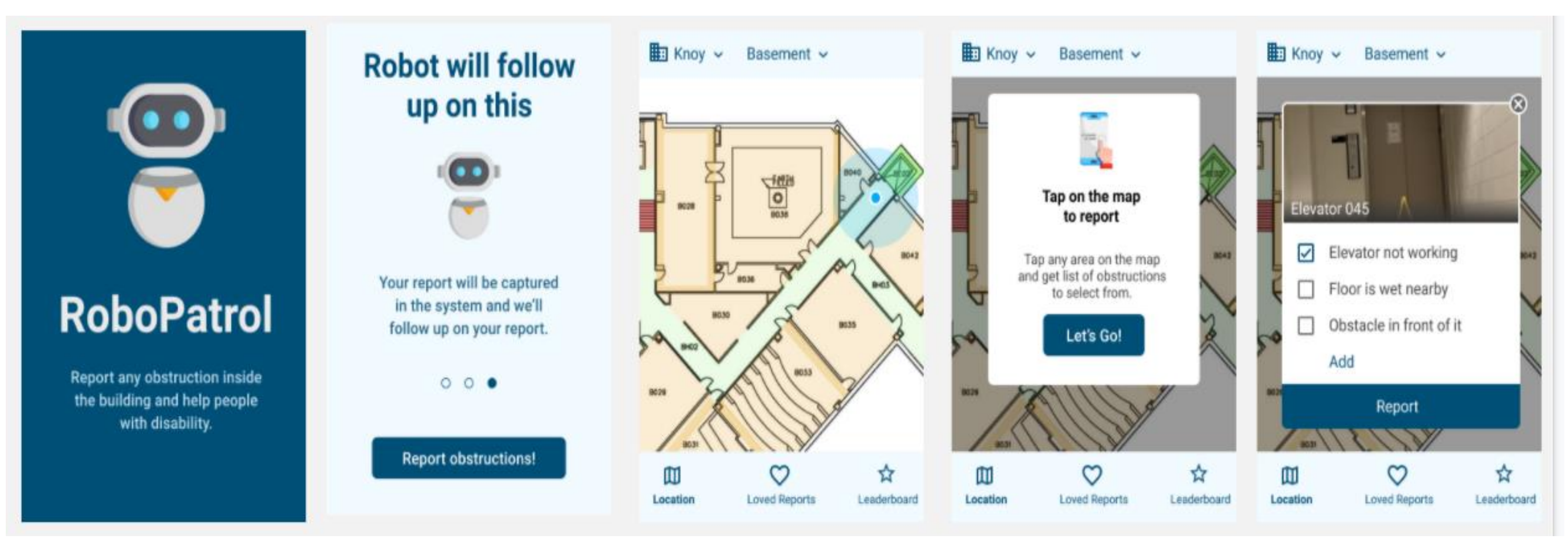}
    \vspace{-5pt}
    \caption{Front-end prototype.} 
    \label{fig:front}
\end{figure}

The system represents the central point that connects all the other sections of the project and consists of two parts: \textit{interface} and \textit{database}. The \textit{interface} is where the participants provide information about obstacles in their location that might affect the navigation of the low vision and visually impaired users. The interface is also where the visually impaired users request and receive information about the safety-levels of their desired path to their indoor destination. Once participants provide information about the presence of an obstacle via the interface, the information is logged in the database and then synced with the robot system to enable the robot to verify the continued existence of the obstacle at the location specified by the participant. The \textit{database} is where obstacle and event information are stored and from where necessary data is transmitted to the robot system for verification. Information in the database is divided into $events$ - which represents information provided by users about ongoing events, $events\_verified$ which represents information about verification of events by the robot. $Obstacles$ contains information about detrimental objects in indoor environments provided by participants and members of the public to the system, $obstacles\_verified$ contains information about obstacles in indoor environments that has been verified by the robot. The database also contains information about registered and verified $users$, $users\_guests$, and $users\_maintenance$, with guests and maintenance representing unregistered users and maintenance staff respectively. This categorization allows for the potential rating of information provided by the different user categories for reliability. The system syncs with the robot operating system using Google Drive API. After the verification of events or obstacles by the robot, the updated information is synced with the database from where the information is then provided to visually impaired with time stamp of when the information was obtained. Figure \ref{fig:image3} highlights how the different components of the system interconnect with each other. Our demo was built with Python and MySQL. And our front-end was a low-fidelity interactive prototype (See Fig. \ref{fig:front} as at the time of this publication.

\subsection{Robot System}
\begin{figure}
    \centering
    \includegraphics[width=0.7\linewidth]{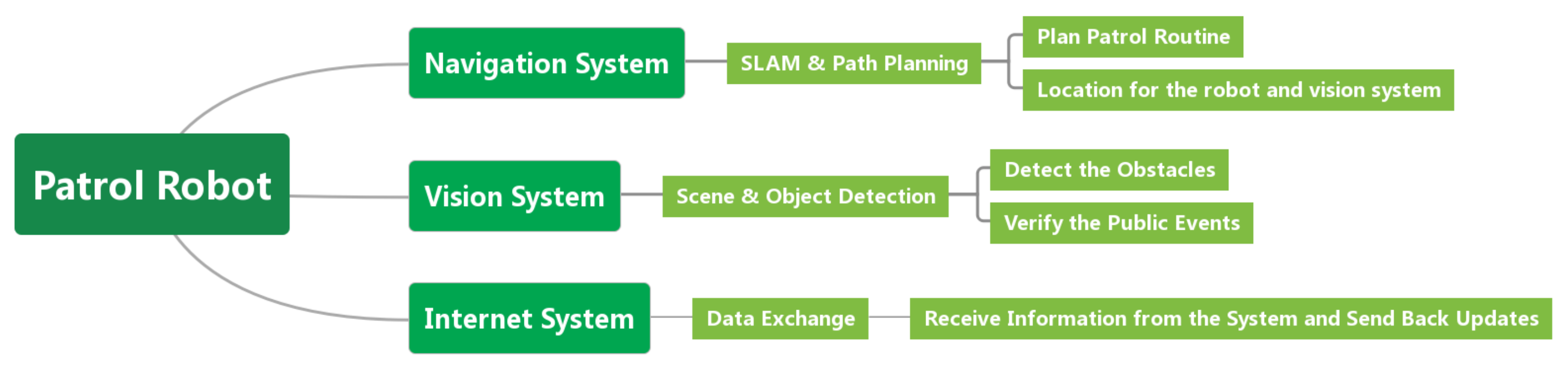}
    \caption{Robot system framework.} 
    \label{fig:robot}
\end{figure}

The patrol robot part consists of three parts: navigation system, vision system, and internet system. (See Fig.  \ref{fig:robot}), and these three systems will be integrated into the Robot Operating System (ROS) ~\cite{quigley2009ros}. The objective of patrol robot system is to verify and update events and obstacles in the environment. We describe these parts in the subsections below.

\subsubsection{Internet System}

The function of internet system is to exchange data with other parts in the assistive system, receiving mission message and sending back updated environment information including events and obstacles. The mission message is stored in a TXT file with the name of Mission Message, sent by Crowd-sourcing database through Google Drive Sync API. The received message consists of the reported events and obstacles which need to be verified: the information of events is the tuple: (\#Event, number, Event keyword, Semantic Location) and the one of obstacles is the tuple: (\#Obstacle, number, Obstacle type, Obstacle number, Semantic Location). The Mission Message file will be overwritten and updated by the Crowd-sourcing database every event. We adapted the NLTK toolkit in Python, to process the text information in the TXT file. The semantic location of event and obstacle is then sent to navigation system, activating the robot to depart for the corresponding check point. The event keyword, obstacle number and type are conveyed to vision system to verify the event and obstacle by comparing with vision recognition results at corresponding check point.
 
Then based on the vision recognition results from vision system and also confirmation of location from the navigation system, the internet system will generate another TXT file with the name of Update, which can be extracted by Crowd-sourcing database through Google Drive Sync API. The file consists of verified event information: (\#Event, number, result), where if the event still exist the robot returns 1 in result, otherwise, 0, and updated obstacle lists: (\#Obstacle, number, Obstacle type, Obstacle number, Semantic Location). The Update file will be overwritten and updated by the internet system every time when a patrol task is completed. Based on the Update file, the crowd-sourcing database system can know if the reported events are still going on and the obstacle level of the current environment, and then update them to the low-vision users.

\subsubsection{Navigation System}
The navigation system aims at providing the patrol path planning and location of the robot. For this goal, we adapt ROS Navigation Package, which provides SLAM and path planning algorithms, based on which the robot will be able to locate itself and plan the patrol routine. 

\begin{figure}
    \centering
    \includegraphics[width=0.5\linewidth]{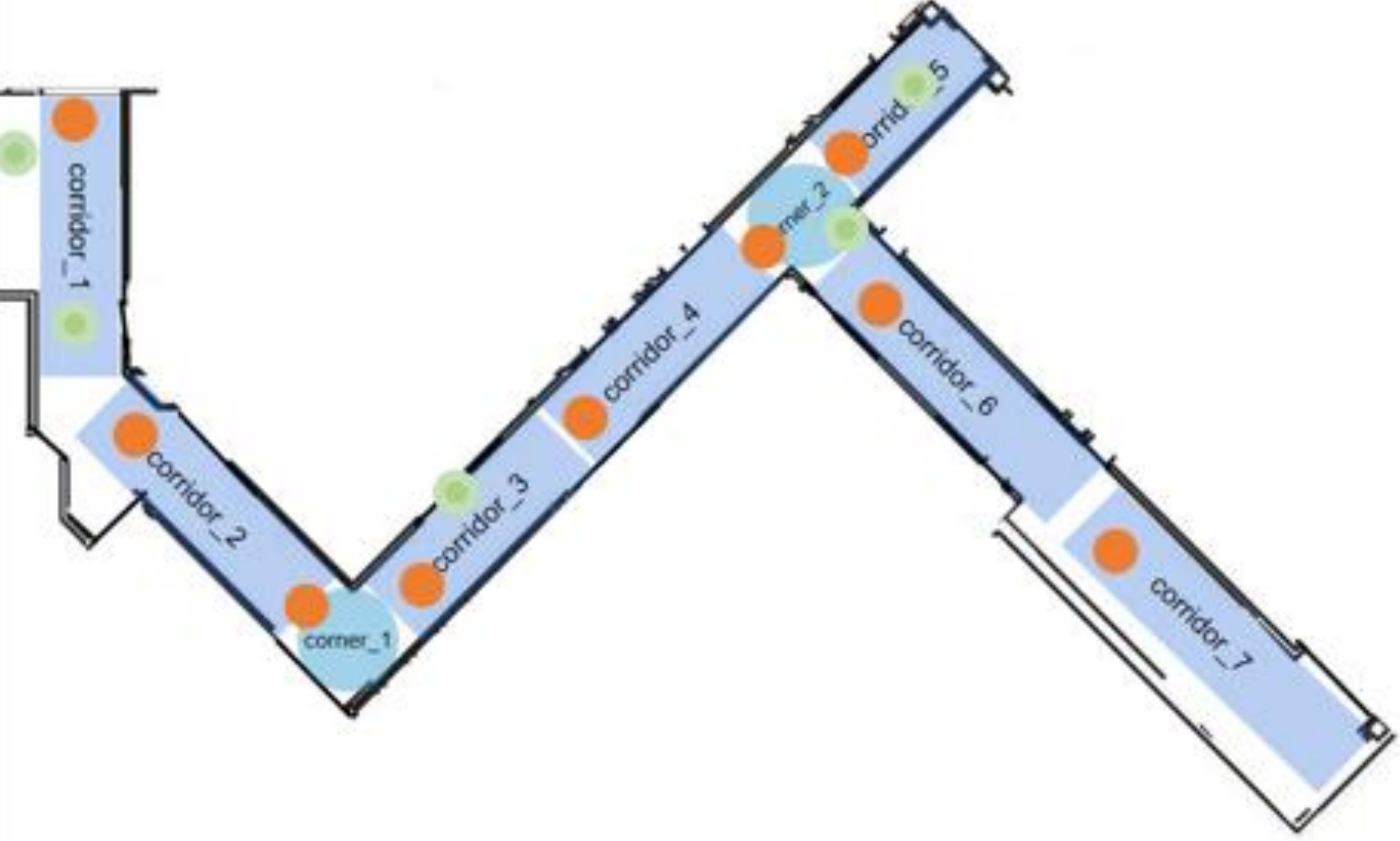}
    \vspace{-8pt}
    \caption{Illustration of semantic areas in robot map.} 
    \label{fig:image9}
\end{figure}

The robot map is separated into several semantic parts based on corridors and corners to make the map syncing between the main system and robot system easier. Each semantic area has one checkpoint, where the robot’s camera can capture a full and clear scenario of the area in order to recognize potential obstacles. In addition to this regular set of checkpoints, we also set another set of checkpoints for event checking, where the robot can see the event scenario clearly. The regular and event checkpoints correspond to $ob\_location$ and $ev\_location$ in the internet system respectively.
For instance, the robot map of experimental site is divided into nine semantic areas and is labeled with nine regular checkpoints (red) and five event checkpoints (green). This semantic map can provide semantic location labels, e.g., $corridor\_5$ or $corner\_2$, for vision system, reflecting the location of the robot and thus the location of recognized obstacles, and the location for robot to check reported events, for example, for the event that elevator \#1 is under maintenance, we set the checkpoint, which is facing the elevator, so that the robot can see the situation of the elevator, e.g., there is a maintenance sign in front of it (See Fig. \ref{fig:image9}).
 
\subsubsection{Vision system}
\begin{figure}
    \centering
    \includegraphics[width=0.7\linewidth]{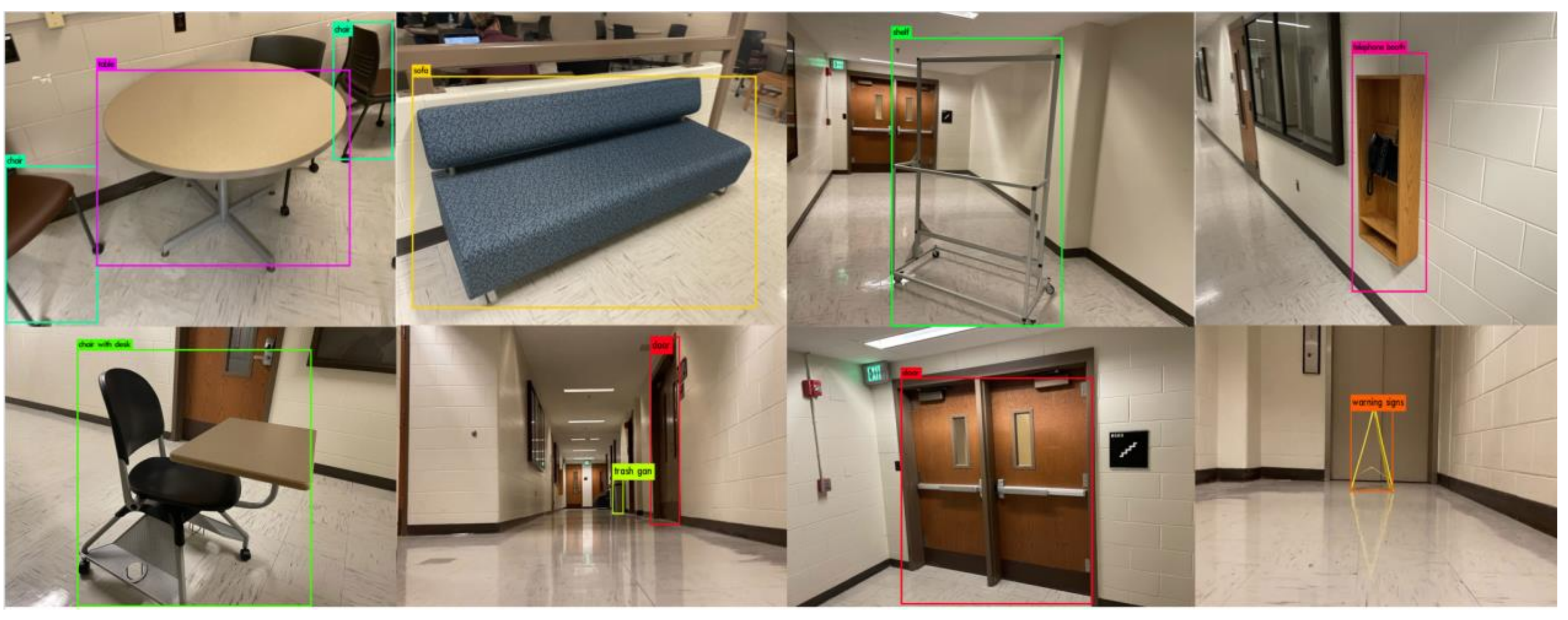}
    \vspace{-5pt}
    \caption{Illustration of recognition results on static images.} 
    \vspace{-5pt}
    \label{fig:image10}
\end{figure}

The vision system oversees the real-time recognition of obstacles and events in the environment. To this end, we utilize YOLO v3 ~\cite{redmon2018yolov3}, which is proven and has a good performance in real-time recognition, as our deep learning model for this exploratory study. 11 kinds of items, which are typical obstacles for low-vision people in school environments, including: table, sofa, chair, telephone booth, trash can, hand-washer rod, warning signal, shelf, chair with desk, people, and door, are regarded as the targets of recognition. We collected 100 sample images for each type to build a database including total 1,100 pictures for training. The trained model achieved 100\% accuracy on testing data set with 550 static images (See Fig. \ref{fig:image10}). However, although the model successfully recognized most of items during real-time testing, it could not achieve 100\% accuracy in real-time recognition due to the light condition, image deforming and object overlapping. We consider adding more samples in different angle and lighting conditions and in overlapping situations to further explore improving accuracy in real-time recognition.

For obstacle update, the robot will reach each regular checkpoint to conduct real-time recognition. After recognition, the model can return the type, number, and bounding box of the recognized objects. Based on the returned information combined with the semantic location labels from navigation system, the vision system can update all the obstacles along the patrol path using a tuple: (Obstacle type, Obstacle number, Semantic Location). For event verification, the robot will get to the event checkpoints mentioned in Mission Message file and perform recognition tasks. For $class\_waiting$ event, if the robot recognizes more than two people at the checkpoint, then it will report the event is still going on and return 1 for $ev\_result$, otherwise, 0. And for the event of $elevator\_repair$, if the robot detects a maintenance sign ahead the elevator, it will return 1 for or $ev\_result$, otherwise, 0. Then after finishing all recognition tasks, the vision system will send the information of event verification and obstacle update back to internet system.

\subsection{Crowdsourcing and Gamification}

\begin{figure}
    \centering
    \includegraphics[width=0.45\linewidth]{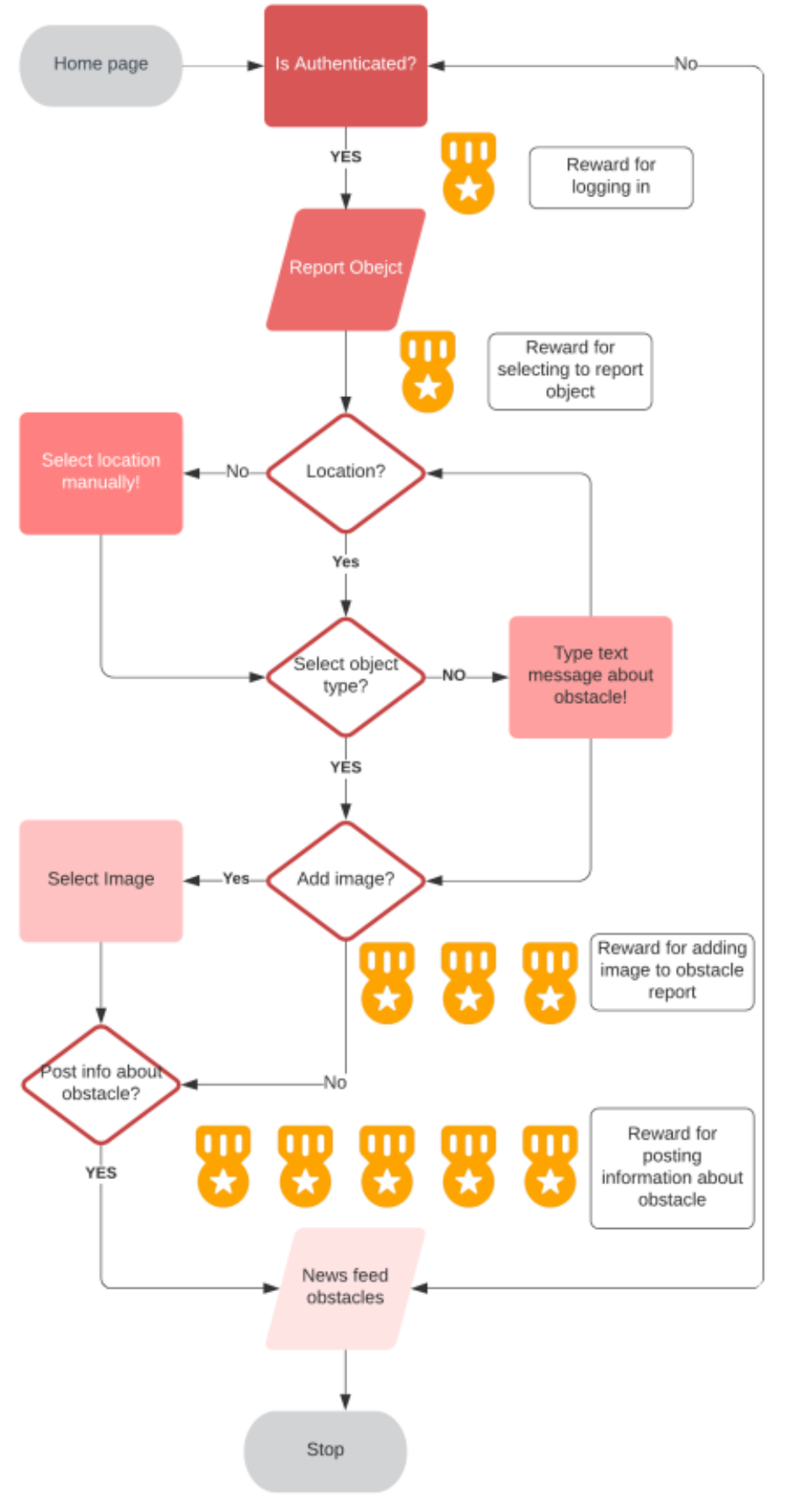}
    \vspace{-5pt}
    \caption{Gamification of the process.} 
    \vspace{-5pt}
    \label{fig:image2}
\end{figure}

Crowdsourcing is a distributed problem-solving methodology that involves using a network of users to complete a task or resolve a problem ~\cite{brabham2013crowdsourcing}. In crowdsourcing, the community of users often work together asynchronously to achieve a predefined goal. We are employing the crowdsourcing framework to recruit, engage, and retain participants who will provide information about the obstacles in an indoor environment to enable the visually impaired users to navigate the space meaningfully and safely. The success of our system relies heavily on the effectiveness of our crowdsourcing and gamification strategies. Hence, our goal involves ensuring that the process of participating is very pleasurable and entertaining to ensure that participants continually engage with the system and also invite people within their local network to participate. To achieve this goal, we inserted gamification and pleasurability loops into our crowdsourcing flowchart as seen in Fig. \ref{fig:image2}. We describe the gamification process below.

\subsubsection{Gamification} 
Games are a great way for people to express their competitive nature ~\cite{hamari2014does}. When embedded within a system as microgames, it motivates people to participate in the events before and after the games so that they can win their prizes. In this context, it will allow participants to gain the double benefit of engaging in an altruistic act and also getting rewarded for helping out. 

\subsubsection{Gamification as an Incentive Mechanism for Sustaining Robot Patrol}
A participant gains a point anytime they log into the app with their user account details. The goal of rewarding the users anytime they log in to the app are two-fold 1) to motivate them to return to the app after they complete their task and 2) to encourage them to participate as registered users instead of guests. Furthermore, if the participant selects the option of reporting an obstacle, they also earn a point. The goal is to make the reporting of an obstacle feel like a real-life game that people typically get rewarded for after completing a task. If the participant completes the process of reporting an obstacle, then they gain 5 points for their efforts. When they complete all the related tasks, they are shown a leaderboard that helps them to visualize the impact of their effort and how it might have helped the visually impaired to navigate indoor spaces meaningfully. They will also be shown the feedback from the visually impaired with their comment of how the information that was provided by the participant helped them to get to their destination meaningfully and safely.  Furthermore, at the end of specified period, the participant with the highest points and badges gets a recognition and reward from the designated authorities for the assistance and contribution to the independence of the visually impaired community.

\subsection{Visually Impaired Users}
The visually impaired users are the main beneficiaries of the information that will be provided by participants and verified by the robot. This information to be sent to the visually impaired will contain contextual guidance, including information about the severity of the obstacle ranging from low, to middle and high. Presenting this severity information contextually instead of numbers reduces the cognitive load that might arise when the information is presented in numbers and forcing the visually to process information in a way that is relatively difficult for them, thus, making the information less meaningful for them. Furthermore, the information is being presented using the audio and haptic modalities to ensure that the visually impaired users have the freedom to determine how they want to consume the information that is being presented to them.

\section{Evaluation and Results}

In this section we focus on limitations and highlight benefits in the concluding section. We conducted a wizard of oz experiment to investigate the benefits and limitations of our approach. This test was conducted across two dimensions - \textit{testing the robot system} and \textit{testing the crowdsourcing framework}. Both tests were conducted in a controlled area that we had secured for the experiment. For \textit{testing the robot system}, we placed multiple obstacles in two different locations within the controlled area and then simulated the process of creating and sending a report about a potential obstacle to the robot system for verification. The robot was then deployed to investigate the obstacles, verify their continued existence, and then report an update back to the system. For \textit{testing the crowdsourcing framework}, one of the researchers for this project played the role of a participant and reported information about a closed elevator to the system. Subsequently, we simulated the role of a visually impaired who has been notified that the elevator has been closed to explore how the visually impaired participants might engage with the information. Conducting this test with visually impaired participants would have been the ideal scenario, however, we employed this testing session as a first-level assessment of our prototype that would allow us to refine the system and then recruit visually impaired users for a more robust testing.

Results from the testing session revealed areas of improvement for the future iteration and scaling of the system. They include 1) \textit{Object recognition challenges:} findings from our experiment revealed that the model could not achieve 100\% accuracy in real-time recognition due to objects overlapping each other, light conditions, and image deforming; 2) \textit{Algorithms:} findings from the testing session revealed the need to explore the suitable algorithms that would be used in determining the location of the participant to allow them to report obstacles more accurately, we had used semantic mapping for this demo, but this would be difficult to scale up; 3) \textit{Quality of information from the public:} there is a need to ensure that accurate and good-intentioned information is being provided to the robot system for verification to ensure efficient use of resources; 4) \textit{Data Privacy:} another concern was about the privacy of the information of the planned destination of the participants; and 5) \textit{test limitation:} there were also concerns about the testing scenario since visually impaired users rarely travel to unfamiliar areas without prior training. In all, we view these findings as learning opportunities that would provoke the research towards new directions.

\section{Conclusion and Future work}
In this project, we developed Robot Patrol, an integrated system that employs a combination of crowdsourcing and robotic frameworks to provide contextual information to the visually impaired to empower them to navigate indoor spaces safely. We conducted an experiment to test the feasibility of the system. Future work involves conducting user-testing with visually impaired users and implementing the platform with a universal design framework based on findings from the testing sessions. That is to say that, although the platform would be developed for the visually impaired, other users should also be able to employ the platform to obtain information about events and obstacles within their vicinity. For example, since the start of the pandemic, facility managers face an increased workload. They have to perform extra cleaning and maintenance of potentially indoor spaces in addition to their traditional role of monitoring and maintaining order. This system can as well support facility managers and provide them with up-to-date information and contextual guidance on spaces that need their attention and the severity of the potential scenario, without them actually having to visit the spaces beforehand, among other potential use cases of the system in indoor environments. In all, this was an insightful project that will continue to evolve.

\begin{acks}
Thanks to the building administrators at our home institution for providing us with construction signs and space for conducting the testing session.
\end{acks}

\bibliographystyle{ACM-Reference-Format}
\bibliography{sample-base}

\appendix

\end{document}